\documentclass[runningheads]{llncs}

\usepackage{eccv}

\usepackage{eccvabbrv}

\usepackage{graphicx}
\usepackage{booktabs}
\usepackage{multirow}
\usepackage{makecell}
\usepackage{pgf-pie} 
\usepackage{pgfplots}

\usepackage[accsupp]{axessibility}  

\usepackage[pagebackref,breaklinks,colorlinks]{hyperref}

\usepackage{orcidlink}

\begin{document}

\title{Pixel-Aware Stable Diffusion for Realistic Image Super-Resolution and Personalized Stylization} 
\titlerunning{Pixel-Aware Stable Diffusion}

\author{Tao Yang\inst{1} \and
Rongyuan Wu\inst{2} \and
Peiran Ren\inst{3} \and
Xuansong Xie\inst{3} \and
Lei Zhang\inst{2}}

\authorrunning{Yang et al.}

\institute{Bytedance Inc. \and
The Hong Kong Polytechnic University \and
Alibaba Group}

\maketitle

\begin{abstract}
Diffusion models have demonstrated impressive performance in various image generation, editing, enhancement and translation tasks. In particular, the pre-trained text-to-image stable diffusion models provide a potential solution to the challenging realistic image super-resolution (Real-ISR) and image stylization problems with their strong generative priors. However, the existing methods along this line often fail to keep faithful pixel-wise image structures. If extra skip connections between the encoder and the decoder of a VAE are used to reproduce details, additional training in image space will be required, limiting the application to tasks in latent space such as image stylization. In this work, we propose a pixel-aware stable diffusion (PASD) network to achieve robust Real-ISR and personalized image stylization. Specifically, a pixel-aware cross attention module is introduced to enable diffusion models perceiving image local structures in pixel-wise level, while a degradation removal module is used to extract degradation insensitive features to guide the diffusion process together with image high level information. An adjustable noise schedule is introduced to further improve the image restoration results. By simply replacing the base diffusion model with a stylized one, PASD can generate diverse stylized images without collecting pairwise training data, and by shifting the base model with an aesthetic one, PASD can bring old photos back to life. Extensive experiments in a variety of image enhancement and stylization tasks demonstrate the effectiveness of our proposed PASD approach. Our source codes are available at \url{https://github.com/yangxy/PASD/}.

\end{abstract}

\section{Introduction}
\label{sec:intro}

Real-world images often suffer from a mixture of complex degradations, such as low resolution, blur, noise, etc., in the acquisition process. While image restoration methods have achieved significant progress, especially in the era of deep learning \cite{dong2014srcnn,lim2017edsr}, they still tend to generate over-smoothed details, partially due to the pursue of image fidelity in the methodology design. By relaxing the constraint on image fidelity, realistic image super-resolution (Real-ISR) aims to reproduce perceptually realistic image details from the degraded observation. The generative adversarial networks (GANs) \cite{goodfellow2014gan} and the adversarial training strategy have been widely used for Real-ISR \cite{ledig2017srgan,wang2018esrgan, zhou2022codeformer, chen2022femasr} and achieved promising results. However, GAN-based Real-ISR methods are still limited in reproducing rich and realistic image details and tend to generate unpleasant visual artifacts. Meanwhile, GAN-based methods have also been widely used in various image stylization tasks such as cartoonization and old-photo restoration. For example, Chen \etal \cite{chen2018cartoongan} proposed CartoonGAN to generate cartoon stylization by using unpaired data for training. However, different models need to be trained for different styles. Wan \etal \cite{wan2021bringing} introduced a triplet domain translation network to restore old photos. While achieving promising results, the multi-stage procedure of this method makes it complex to use.

\begin{figure}[t!]
    \centering
    \begin{subfigure}{0.3\textwidth}
    \includegraphics[width=0.9\linewidth]{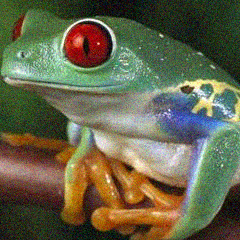}
    \caption{LQ}
    \end{subfigure}
    \begin{subfigure}{0.3\textwidth}
    \includegraphics[width=0.9\linewidth]{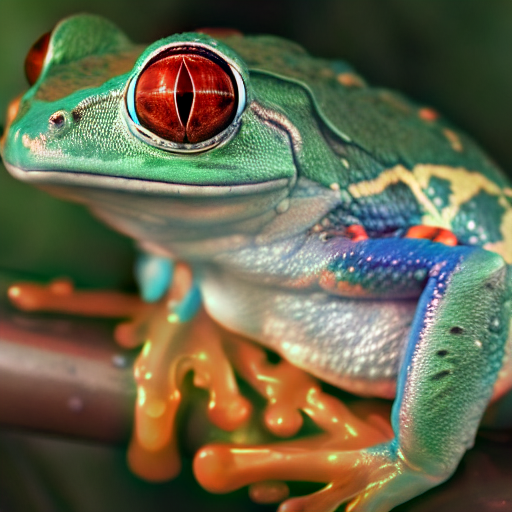}
    \caption{ControlNet}
    \end{subfigure}
    \begin{subfigure}{0.3\textwidth}
    \includegraphics[width=0.9\linewidth]{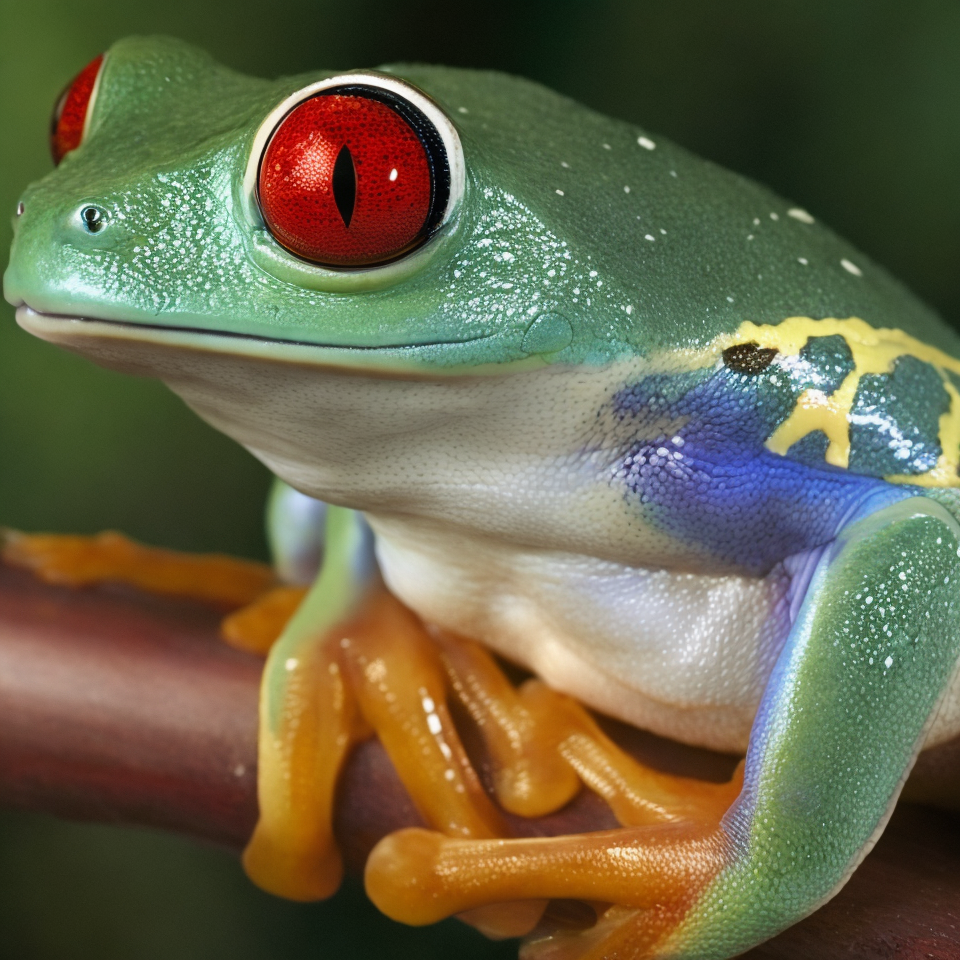}
    \caption{PASD}
    \end{subfigure}
    \vspace*{-2mm}
    \caption{From left to right: an input LQ image, the Real-ISR outputs by ControlNet \cite{zhang2023controlnet} and our PASD method. One can see that the output by ControlNet has clear content inconsistency with the input, while PASD preserves the structure well in pixel level.}
    \label{fig:example}
    \vspace*{-3mm}
\end{figure}

Recently, denoising diffusion probabilistic models (DDPMs) have shown outstanding performance in tasks of image generation \cite{ho2020ddpm}, and it has become a strong alternative to GAN due to its powerful capability in approximating diverse and complicated distributions. With DDPM, the  pre-trained text-to-image (T2I) and text-to-video (T2V) latent diffusion models \cite{rombach2021latent,ramesh2022dalle2,saharia2022imagen,ho2022imagenvideo} have been popularly used in numerous downstream tasks, including personalized image generation \cite{ruiz2023dreambooth,kumari2022customdiffusion}, image editing \cite{brooks2022instructpix2pix,kawar2023imagic}, image inpainting \cite{yang2022paintbyexample} and conditional image synthesis \cite{zhang2023controlnet}. Diffusion models have also been adopted to solve image restoration tasks. A denoising diffusion restoration model (DDRM) is proposed in \cite{kawar2022ddrm} to solve inverse problem by taking advantage of a pre-trained denoising diffusion generative model. However, DDRM assumes a linear image degradation model, limiting its application to more practical scenarios such as Real-ISR. 
Considering that the pre-trained T2I models such as Stable Diffusion (SD) \cite{rombach2021latent} can generate high-quality natural images, Zhang and Agrawala \cite{zhang2023controlnet} proposed ControlNet, which enables conditional inputs like edge maps, segmentation maps, etc., and demonstrated that the generative diffusion priors are also powerful in conditional image synthesis. Unfortunately, ControlNet is not suitable for pixel-wise conditional control (see Fig.~\ref{fig:example} for an example). Qin \etal ~\cite{qin2023unicontrol} extended ControlNet by introducing UniControl to enable more diverse visual conditions. Liu \etal  ~\cite{liu2023colorization} and Wang \etal  ~\cite{wang2023stablesr} demonstrated that pre-trained SD priors can be employed for image colorization and Real-ISR, respectively. However, they resorted to a skipped connection to pass pixel-level details for image restoration, requiring extra training in image space and limiting the model capability to tasks performed in latent space such as image stylization.
 
In this work, we aim to develop a flexible model to achieve Real-ISR and personalized stylization by using pre-trained T2I models such as SD \cite{rombach2021latent}, targeting at reconstructing photo-realistic pixel-level structures and textures. Our idea is to introduce pixel-aware conditional control into the diffusion process so that robust and perceptually realistic outputs can be achieved. To this end, we present a pixel-aware cross attention (PACA) module to perceive pixel-level information without using any skipped connections. A degradation removal module is employed to reduce the impact of unknown image degradations, alleviating the burden of diffusion module to handle real-world low-quality images. We also demonstrate that the high-level classification/detection/captioning information extracted from the input image can further boost the Real-ISR performance. Inspired by recent works \cite{lin2023snr,girdhar2023emuvideo}, we present an adjustable noise schedule to further boost the performance of Real-ISR and image stylization tasks. In particular, the proposed method, namely pixel-aware stable diffusion (PASD), can perform personalized stylization tasks (\eg, caroonization and old photo restoration) by simply shifting the base model to a personalized one. Extensive experiments demonstrate the effectiveness and flexibility of PASD. 

\section{Related Work}
\textbf{Realistic Image Super-Resolution.}
Though deep learning based image super-resolution \cite{dong2014srcnn,lim2017edsr} has achieved significant progress, they still suffer from over-smoothed details due to the high illness of the task by minimizing the fidelity objectives (\eg, PSNR, SSIM). Realistic image super-resolution (Real-ISR) aims to reproduce perceptually photo-realistic image details by optimizing not only the fidelity but also the perception objectives. The GAN \cite{goodfellow2014gan} network and its adversarial training strategies are widely in Real-ISR \cite{ledig2017srgan,wang2018esrgan}. Basically, a generator network is used to reconstruct the desired high-quality (HQ) image from the low-quality (LQ) input, while a discriminator network is used to judge whether the HQ output is perceptually realistic. In the early study, bicubic downsampling or some simple degradations \cite{dong2014srcnn,lai2017lapsrn} are used to simulate the LQ-HQ training pairs. Cai \etal ~\cite{cai2019realsr} collected a real-world dataset with paired LQ-HQ images by zooming camera lens. Zhang \etal ~\cite{zhang2021bsrgan} and Wang \etal ~\cite{wang2021realesrgan} later modeled complex degradations by shuffling degradation types and using a high-order process, respectively.  
Recently, Yang \etal ~\cite{yang2023syn} took the advantages of handcrafted degradation models and generative diffusion models to synthesize realistic LQ-HQ training pairs.

Though GAN-based models have dominated the previous research in Real-ISR, adversarial training is unstable and the GAN-based Real-ISR methods often bring unnatural visual artifacts. Liang \etal ~\cite{liang2022ldl} proposed a locally discriminative learning approach to suppress the GAN-generated artifacts, yet it is difficult to introduce additional details. Recently, inspired by the success of generative priors in face restoration tasks \cite{yang2021gpen,wang2021gfpgan}, some works have been proposed to leverage the priors learned by diffusion models \cite{ho2020ddpm} and pre-trained T2I models \cite{rombach2021latent} to solve the Real-ISR problems and obtain interesting results \cite{chen2022femasr,kawar2022ddrm,wang2023stablesr,wu2023seesr}. In this work, we aim to develop an SD based Real-ISR model which can achieve pixel-level restoration of image details and textures.  

\textbf{Personalized Stylization.}
Inspired by the powerful learning capacity of deep neural networks, Gatys \etal ~\cite{gatys2015style} presented an optimization based method to transfer the style of a given artwork to a content image. This work was extended by many following researches \cite{johnson2016perceptual,li2017WCT,zhang2023inst}. However, all these methods require an extra image as style input. This problem can be alleviated by resorting to an image-to-image framework \cite{zhu2017cyclegan,chen2018cartoongan,chen2020animegan}. Due to the lack of pairwise training data, some works \cite{men2022dctnet,yang2022beyond} focus on portrait stylization with the help of StyleGAN \cite{karras2019stylegan}. With the rapid development of SD models \cite{rombach2021latent}, some works \cite{brooks2022instructpix2pix,zhang2023controlnet} generate stylized images by using proper instruction prompts, achieving impressive results. However, these methods fail to maintain pixel-wise image structures in the stylization process, and they lack the ability to mimic the appearance of subjects in a given reference set. To meet the specific needs of different users, Ruiz \etal ~\cite{ruiz2023dreambooth} and Kumari \etal ~\cite{kumari2022customdiffusion} proposed personalized stylization approaches for T2I diffusion models. While achieving pleasant stylization results, these methods require extra training procedures for different reference sets.


Old photo restoration, which is a challenging task due to the unknown mixed degradations, can also be viewed as a stylization problem. Most existing works focus on a specific issue such as crack removal \cite{yu2018deepfill}, super-resolution \cite{wang2021realesrgan,chen2022femasr} and facial region enhancement \cite{wang2021gfpgan,yang2021gpen}. Wan \etal \cite{wan2021bringing} proposed to address this problem using a novel triplet domain translation network. They narrowed the domain gap between real old photos and synthetic ones in the compact latent space and learned to restore old photos via latent space translation. This multi-stage approach yields interesting, yet it is unstable and is not easy to use.

\textbf{Diffusion Probabilistic Models.}
The seminal work of DDPM \cite{ho2020ddpm} has demonstrated strong capability in generating high quality natural images. Considering that DDPMs require hundreds of sampling steps in the denoising process, Song \etal ~\cite{song2021ddim} proposed DDIM to accelerate the sampling speed. Following works extend DDPM/DDIM by adapting high-order solvers \cite{lu2022dpm} and distillations \cite{meng2023distill}.
Rombach \etal ~\cite{rombach2022latent} extended DDPM to latent space and demonstrated impressive results with less computational costs. This work sparks the prosperity of large pre-trained T2I and T2V diffusion models such as SD \cite{rombach2021latent}, Imagen \cite{ho2022imagenvideo}.
It has been demonstrated that T2I diffusion priors are more powerful than GAN priors in handling diverse natural images \cite{rombach2021latent,ramesh2022dalle2,saharia2022imagen}. Kawar \etal ~\cite{kawar2023imagic} applied complex text-guided semantic editing to real images. ControlNet \cite{zhang2023controlnet} enables conditional inputs, such as edge maps, segmentation maps, keypoints, \etc, to T2I models. Liu \etal ~\cite{liu2023colorization} and Wang \etal ~\cite{wang2023stablesr} respectively utilized generative diffusion priors to image colorization and super-resolution.

\section{Pixel-Aware Stable Diffusion Network}

\begin{figure*}[t!]
    \centering
    \includegraphics[width=0.96\textwidth]{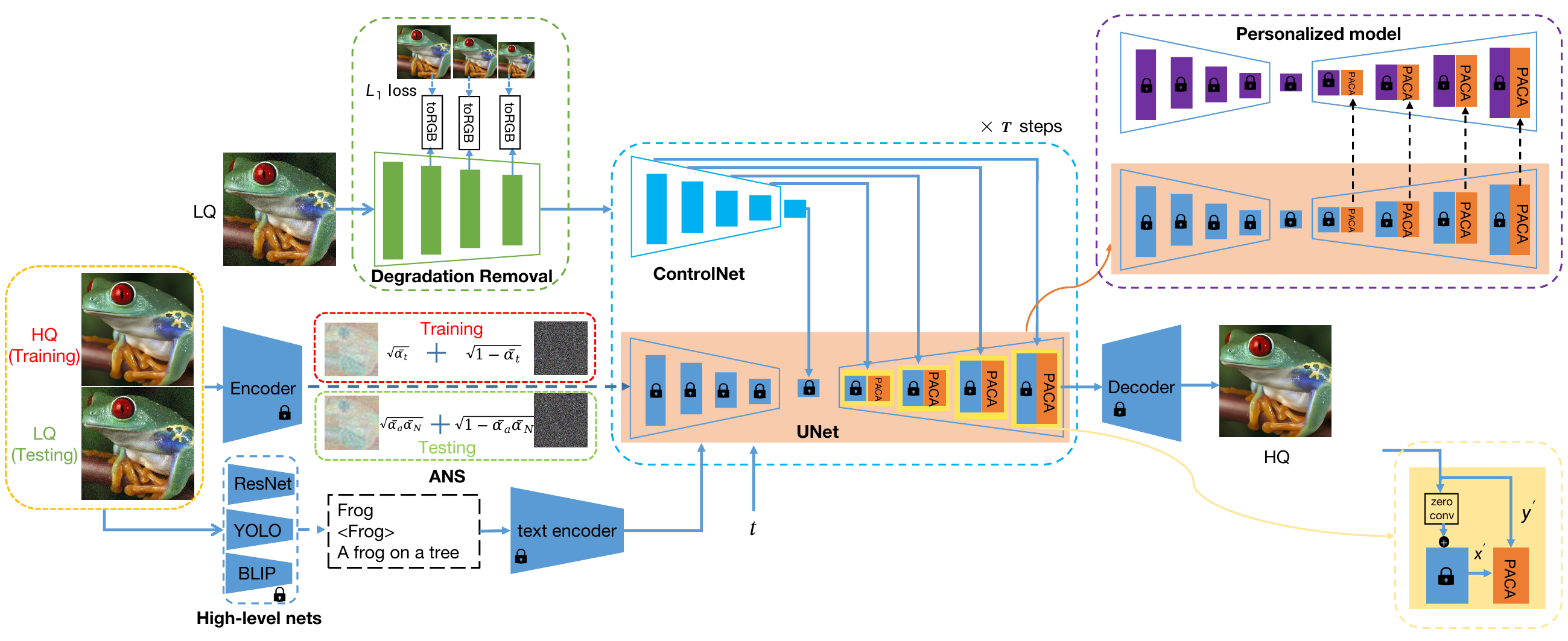}
    \vspace*{-1mm}
    \caption{Architecture of the proposed pixel-aware stable diffusion (PASD) network. PASD consists of several modules, including Degradation Removal, ControlNet, PACA, ANS, and High-level Nets. During training, the encoder maps the HQ image to a latent representation, which is then added by noise to yield a noisy latent. In testing, the LQ image is used to generate the noisy input, while ANS is employed for flexible perception-fidelity balance. The noisy latent is fed to the UNet along with timestep, high-level information, and the output of ControlNet after Degradation Removal conditioned on the LQ image. In particular, the output of ControlNet is added to the UNet via PACA in latent space. PASD can be readily used for personalized stylization by simply switching the base diffusion model to a personalized one.}
    \label{fig:arch}
    \vspace*{-2mm}
\end{figure*}

Our method is based on generative diffusion priors. In particular, we utilize the powerful pre-trained SD \cite{rombach2021latent} model, while alternative diffusion models such as DALLE2 \cite{ramesh2022dalle2} and Imagen \cite{saharia2022imagen} can also be adopted. The architecture of our pixel-aware stable diffusion (PASD) network is depicted in Fig.~\ref{fig:arch}. One can see that in addition to the pre-trained SD model, PASD has four main modules: a degradation removal module to extract degradation insensitive low-level control features, a high-level information extraction module to extract semantic control features, an adjustable noise schedule (ANS) and a pixel-aware cross-attention (PACA) module to perform pixel-level guidance for diffusion. In addition to the Real-ISR task, our PASD can be readily used for personalized stylization by simply switching the base diffusion model to a personalized one.     

\subsection{Degradation Removal Module}
Real-world LQ images usually suffer from complex and unknown degradations. We thus employ a degradation removal module to reduce the impact of degradations and extract ``clean'' features from the LQ image to control  the diffusion process. As shown in Fig.~\ref{fig:arch}, we adopt a pyramid network to extract multi-scale feature maps with 1/2, 1/4 and 1/8 scaled resolutions of the input LQ image.
Intuitively, it is anticipated that these features can be used to approximate the HQ image at the corresponding scale as close as possible so that the subsequent diffusion module could focus on recovering realistic image details, alleviating the burden of distinguishing image degradations. Therefore, we introduce an intermediate supervision by employing a convolution layer ``toRGB'' to turn every single-scale feature maps into the HQ RGB image space. We apply an $L_1$ loss on each resolution scale to force the reconstruction at that scale to be close to the pyramid decomposition of the HQ image: 
$\mathcal{L_{DR}}=\sum_s ||\mathbf{I}_{hq}^{s}-\mathbf{I}_{sr}^s||_1$,
where $\mathbf{I}_{hq}^{s}$ and $\mathbf{I}_{sr}^s$ represent the HQ ground-truth and ISR output at scale $s$. Note that this module is only required in the Real-ISR task. 

\subsection{Pixel-Aware Cross Attention (PACA)}
The main challenge of utilizing pre-trained T2I diffusion priors for image restoration tasks lies in how to enable the diffusion process be aware of image details and textures in pixel-level. The well-known ControlNet can support task-specific conditions (\eg, edges, segmentation masks) well but fail for pixel-level control. Given a feature map $\textbf{x}\in\mathbb{R}^{h\times w\times c}$  from U-Net, where $\{h,w,c\}$ are feature height, width and channel numbers, and a skipped feature map $\textbf{y}\in\mathbb{R}^{h\times w\times c}$ from ControlNet, Zhang and Agrawala \cite{zhang2023controlnet} proposed a unique type of convolution layer $\mathcal{Z}$ called ``zero convolution'' to connect them:
\begin{equation}
   \tilde{\mathbf{x}}=\mathbf{x}+\mathcal{Z}(\mathbf{y}),
\end{equation}
where $\tilde{\mathbf{x}}$ is the output feature map. The zero convolution is easy-to-implement. However, simply adding the feature maps from the two networks may fail to pass pixel-level precise information, leading to structure inconsistency between the input LQ and output HQ images. Fig.~\ref{fig:example} shows an example. One can see that by simply applying ControlNet to the LQ input, there are obvious structure inconsistencies in the output image by ControlNet.

To address this problem, some methods employ a skipped connection outside the U-Net \cite{wang2023stablesr} to add image details. However, this introduces additional training in image feature domain, and limits the application of the trained network to tasks performed in latent space (\eg, image stylization). In this work, we introduce a simple pixel-aware cross attention (PACA) to solve this issue. We reshape $\mathbf{x}$ and $\mathbf{y}$ to $\mathbf{x}'\in\mathbb{R}^{h*w\times c}$ and $\mathbf{y}'\in\mathbb{R}^{h*w\times c}$, and consider $\mathbf{y}'$ as the context input. The PACA (see the brown-colored block in Fig.~\ref{fig:arch}) can be computed as follows:
\begin{equation}
   PACA(\mathbf{Q},\mathbf{K},\mathbf{V}) = Softmax(\frac{\mathbf{Q}\mathbf{K}^T}{\sqrt{d}})\cdot\mathbf{V},
\end{equation}
where $\mathbf{Q}$, $\mathbf{K}$, $\mathbf{V}$ are calculated according to operations $to\_q(\mathbf{x}'), to\_k(\mathbf{y}')$ and $to\_v(\mathbf{y}')$, respectively. 

The conditional feature input $\mathbf{y}'$ is of length $h*w$, which equals to the total number of pixels of latent feature $\mathbf{x}$. Since feature $\mathbf{y}'$ has not been converted into the latent space by the Encoder, it preserves well the original image structures. Therefore, our PASD model can manage to perceive pixel-wise information from the conditional input $\mathbf{y}'$ via PACA. As can be seen in the experimental result section, with the help of PACA, the output of our PASD network can reproduce realistic and faithful image structures and textures in pixel-level.

\subsection{Adjustable Noise Schedule (ANS)}
\label{sec:schedule}
As discussed in previous works \cite{lin2023snr,girdhar2023emuvideo}, the noise schedule used in SD \cite{rombach2021latent} suffers train-test discrepancy. In training, the noise schedule leaves some residual signal even at the terminal diffusion timestep $N$, leading to non-zero signal-to-noise ratio (SNR). This weakens the model performance at test time when we sample from random Gaussian noise without the signal information. To address this issue, we propose an adjustable noise schedule (ANS) by introducing signal information from the input image at test time. 

The residual signals at training stage are from the HQ ground-truth data, which are unavailable at test time. In applications such as Real-ISR, we can embed the input LQ latent into the initial random Gaussian noise at terminal diffusion timestep $N$ as a compensation:
\begin{equation}
\mathbf{z}_N=\sqrt{\bar{\alpha}_N}\mathbf{z}_{LR}+\sqrt{1-\bar{\alpha}_N}\mathbf{z},
\end{equation}
where $\mathbf{z}_N$, $\mathbf{z}_{LR}$, $\bar{\alpha}_N$, $\mathbf{z}$ are respectively the latent input at timestep $N$, the LQ latent, the cumulative product of $\alpha$, and the initial random Gaussian noise. This remedy can partially alleviate the discrepancy issue and has been adopted in \cite{wang2023stablesr, wu2023seesr}. However, the train-test discrepancy still exists due to the different origins of residual signals, which can harm the restoration results when the LQ image suffers from severe degradations.

To suppress the side effect of residual signal from LQ data, we introduce an additional Gaussian noise $\mathbf{z}'$ with level $\bar{\alpha}_a\in[0,1]$ into Eq. (3) as:
$\mathbf{z}_N=\sqrt{\bar{\alpha}_a}(\sqrt{\bar{\alpha}_N}\mathbf{z}_{LR}+\sqrt{1-\bar{\alpha}_N}\mathbf{z})+\sqrt{1-\bar{\alpha}_a}\mathbf{z}'=\sqrt{\bar{\alpha}_a\bar{\alpha}_N}\mathbf{z}_{LR}+\sqrt{\bar{\alpha}_a-\bar{\alpha}_a\bar{\alpha}_N}\mathbf{z}+\sqrt{1-\bar{\alpha}_a}\mathbf{z}'$.
Since Gaussian noises $\mathbf{z}$ and $\mathbf{z}'$ are independent, the combination of them is equivalent to another Gaussian noise $\mathbf{z}''$. The above formula can be re-written as follows:
\begin{align}
\mathbf{z}_N&=\sqrt{\bar{\alpha}_a\bar{\alpha}_N}\mathbf{z}_{LR}+\sqrt{1-\bar{\alpha}_a\bar{\alpha}_N}\mathbf{z}''.
\label{eqn:ans}
\end{align}
In this way, by choosing a proper value of $\bar{\alpha}_a$, we can adjust the strength of the residual signal $\mathbf{z}_{LR}$ to enable flexible perception-fidelity trade-off. 



\subsection{High-Level Information}
Our method is based on the pre-trained SD model where text is used as the input, while in tasks such as Real-ISR, the LQ image is available as the input. Though some SD-based Real-ISR methods \cite{wang2023stablesr} adopt the null-text prompt, it has been demonstrated that content-related captions could improve the synthesis results \cite{rombach2021latent}. As shown in Fig.~\ref{fig:arch}, we employ the pre-trained ResNet \cite{he2016resnet}, YOLO \cite{redmon2016yolo} and BLIP \cite{li2023blip2} networks to extract image classification, object detection and image caption information from the LQ input, and employ the CLIP \cite{radford2021clip} encoder to convert the text information into image-level features, providing additional semantic signal to control the diffusion process. 

The classifier-free guidance \cite{ho2021cfg} is adopted in our method:
   $\tilde{\mathbf{\epsilon}}(\mathbf{z}_t,\mathbf{c})=\mathbf{\epsilon}(\mathbf{z}_t,\mathbf{c})+\omega \mathbf{\epsilon}(\mathbf{z}_t,\mathbf{c}_{neg})$,
where $\tilde{\mathbf{\epsilon}}(\mathbf{z}_t,\mathbf{c})$ and $\mathbf{\epsilon}(\mathbf{z}_t,\mathbf{c}_{neg})$ are conditional and unconditional $\mathbf{\epsilon}$-predictions \cite{ho2020ddpm}, $\mathbf{c}$ and $\mathbf{c}_n$ are respectively the positive and negative text prompts, $\mathbf{z}_t$ is the latent feature at step $t$, and $\omega$ adjusts the guidance scale. The unconditional $\mathbf{\epsilon}$-prediction $\mathbf{\epsilon}(\mathbf{z}_t,\mathbf{c}_{neg})$ can be achieved with negative prompts. In practice, we empirically combine words like ``noisy'', ``blurry'', ``low resolution'' as negative prompts, which play a key role to trade off mode coverage and sample quality during inference. It is optional but could boost much the Real-ISR performance.

\subsection{Application to Personalized Stylization}

\textbf{Personalized stylization.}
Thanks to the open source of SD \cite{rombach2021latent} and the recently developed techniques such as DreamBooth \cite{ruiz2023dreambooth} and LORA \cite{hu2022lora}, the community becomes highly prosperous. Contributors can upload a large amount of personalized models finetuned on SD with self-collected data. Since PASD is based on pretrained SD model and the pretrained weights are frozen during model training, it is easy to replace the base model with personalized ones at test time \cite{guo2023animatediff} (as illustrated in the top-right corner of Fig.~\ref{fig:arch}) so that PASD can re-target the output domain and produce stylized results. 

Unlike previous methods \cite{zhu2017cyclegan,chen2018cartoongan,chen2020animegan} that achieve stylization ability by learning a pixel-to-pixel mapping function using adversarial training, our PASD approach decouples stylization generation and pixel-to-pixel mapping, opening a new door for image stylization. By fine-tuning personalized SD models with a batch of style images or downloading different personalized models from online communities \footnote{\url{https://civitai.com/}; \url{https://huggingface.co/}}, one can easily generate various stylized results with our PASD method. In this paper, we use cartoonization as a typical stylization task in experiments.  


\textbf{Old photo restoration.}
Apart from cartoonization, another popular family of personalized models are the aesthetic ones, \ie, those trained on images with a particular aesthetic taste. One typical task of this kind is old photo restoration. By replacing the base model with an aesthetic one, PASD can improve the quality and aesthetics of the input old photo image simultaneously, as will be demonstrated in our experiments.

\subsection{Training Strategy}
\label{sec:training}
In the model training, we first obtain the latent representation $\mathbf{z}_0$ of an HQ image, and progressively add noise to it to yield a noisy latent $\mathbf{z}_t$, where $t$ is a randomly sampled diffusion step. Given a number of conditions such as diffusion step $t$, LQ input $\mathbf{I}_{lq}$ and text prompt $\mathbf{c}$, we learn a PASD network $\mathbf{\epsilon}_\theta$ to predict the noise added to the noisy latent $\mathbf{z}_t$. The optimization objective is: 
\begin{equation}
   \mathcal{L_{DF-\epsilon}} = \mathbb{E}_{\mathbf{z}_0,t,\mathbf{c},\mathbf{I}_{lq},\mathbf{\epsilon}\sim\mathcal{N}(0,1)}\big{[}||\mathbf{\epsilon}-\mathbf{\epsilon}_\theta(\mathbf{z}_t,t,\mathbf{c},\mathbf{I}_{lq})||^2_2\big{]}.
\end{equation}
During the training of Real-ISR models, we jointly update the degradation removal module. The total loss is
    $\mathcal{L} = \mathcal{L_{DF-\epsilon}} + \gamma\mathcal{L_{DR}}$,
where $\gamma$ is a balancing parameter. We simply set $\gamma=1$ in the experiments.
We freeze all the parameters in pre-trained SD, and only train the newly added modules, including the degradation removal module, ControlNet and PACA. The employed ResNet, YOLO and BLIP and CLIP networks for high-level information extraction are also fixed. During training, we randomly replace $50\%$ of the text prompts with null-text prompts. This encourages our PASD model to perceive semantic contents from input LQ images as a replacement of text prompts. 


\section{Experiments}
\label{sec:exp}
\subsection{Experiment Setup}
We adopt the Adam optimizer \cite{kingma2015adam} to train PASD with a batch size of $4$. The learning rate is fixed as $5\times 10^{-5}$. The model is updated for $500K$ iterations with $8$ NVIDIA Tesla 32G-V100 GPUs.

\textbf{Training and testing datasets.}
For the Real-ISR task, we train PASD on DIV2K \cite{timofte2017div2k}, Flickr2K \cite{agustsson2017flicker2k}, OST \cite{wang2018ost}, and the first $10,000$ face images from FFHQ \cite{karras2019stylegan}. We employ the degradation pipeline of Real-ESRGAN \cite{wang2021realesrgan} to synthesize LQ-HQ training pairs. We evaluate our approach on both synthetic and real-world datasets. The synthetic dataset is generated from the DIV2K validation set following the Real-ESRGAN degradation pipeline. For real-world test dataset, we use the RealSR \cite{cai2019realsr} and DRealSR \cite{wei2020cdc} for evaluation.

For the task of cartoonization, we simply reuse the PASD model trained for Real-ISR task and shift the base model with stylized ones obtained from online communities. We conduct comparisons on the first $100$ face images from FFHQ as well as the first $100$ images from Flicker2K.


For the task of old photo restoration, we also adopt the pre-trained PASD model in the task of Real-ISR. Unlike cartoonization, we replace the base model with aesthetic ones. We collect $100$ old photos from Internet for testing.



\textbf{Evaluation metrics.}
For quantitative evaluation of Real-ISR models, we employ the widely used perceptual metrics, including FID \cite{heusel2017fid}, LPIPS \cite{zhang2018lpips}, DISTS \cite{ding2020dists}, NIQE \cite{mittal2013niqe}, MUSIQ \cite{ke2021musiq} and QAlign \cite{wu2023qalign}, to compare the competing Real-ISR models. 
The PSN and SSIM indices (evaluated on the Y channel in YCbCr space) are also reported for reference only because they are not suitable to evaluate generative models. 
For the tasks of cartoonization and old photo restoration, we employ FID \cite{heusel2017fid}, MUSIQ \cite{ke2021musiq} and QAlign \cite{wu2023qalign} for evaluation since the ground-truth images are unavailable.


In addition, for all tasks we invite $15$ volunteers to conduct a user study on $40$ real-world images. Each volunteer is asked to choose the most preferred one among  the outputs of all competing methods, which are presented to the volunteers in random order. 

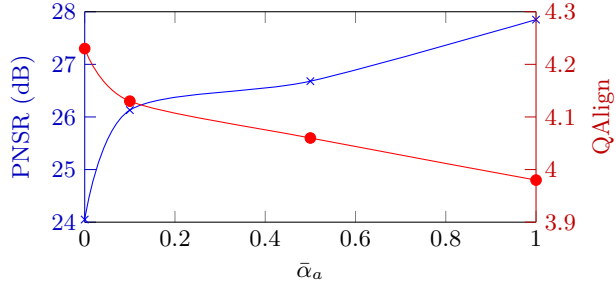
\begin{figure}
    \centering
    \begin{tikzpicture}
    \pgfplotsset{
        scale only axis,
        xmin=0, xmax=1,
        y axis style/.style={
            yticklabel style=#1,
            ylabel style=#1,
            y axis line style=#1,
            ytick style=#1
       },
       width=6cm, height=2.8cm, compat=1.3,
    }
    
    \begin{axis}[
      axis y line*=left,
      y axis style=blue!75!black,
      ymin=24, ymax=28,
      xlabel=$\bar{\alpha}_a$,
      ylabel=PNSR (dB),
    ]
    \addplot[smooth,mark=x,blue] 
      coordinates{
        (0,24.05)
        (0.1,26.13) 
        (0.5,26.68)
        (1,27.85)
    };
    \end{axis}
    
    \begin{axis}[
      axis y line*=right,
      axis x line=none,
      ymin=3.9, ymax=4.3,
      ylabel=QAlign,
      y axis style=red!75!black
    ]
    \addplot[smooth,mark=*,red] 
      coordinates{
        (0,4.23)
        (0.1,4.13) 
        (0.5,4.06)
        (1,3.98)
    };
    \end{axis}
    \end{tikzpicture}
    \vspace{-3mm}
    \caption{The curves of PSNR/QAlign versus $\bar{\alpha}_a$.}
    \label{fig:curve}
    \vspace{-2mm}
\end{figure}

\begin{figure}
  \centering
  \includegraphics[width=.96\linewidth]{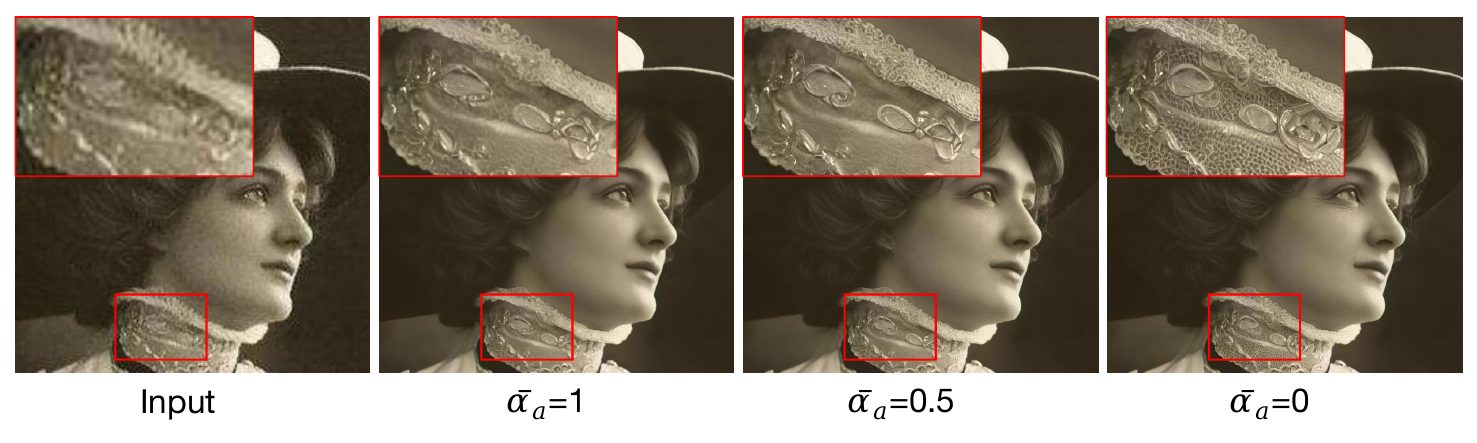}
  \vspace{-2mm}
  \caption{Visual results with different values of $\bar{\alpha}_a$.}
  \label{fig:ans}
  \vspace{-2mm}
\end{figure}

\begin{table*}
    \centering
    \caption{The PSNR, SSIM, LPIPS, FID, DISTS, MUSIQ, NIQE and QAlign indices of different Real-ISR models on synthesized (DIV2K) and real-world (RealSR and DRealSR) test datasets. The best and second best scores are in bold and underscore.}
    \vspace*{-1mm}
        \resizebox{1.\textwidth}{!}{\begin{tabular}{c|c|c c c|c|c c c|c}
            Datasets & Metrics & RealESRGAN & FeMaSR & SwinIR & ResShift & StableSR & DiffBIR & SeeSR & PASD \\
          \hline\hline
            \multirow{8}{*}{\makecell{DIV2K \\ valid}} & PSNR$\uparrow$ & \underline{21.9697} & 21.0272 & 21.6136 & 21.8165 & \textbf{22.3377} & 21.2447 & 21.7714 & 21.6494 \\
            & {SSIM$\uparrow$} & \textbf{0.6164} & 0.5543 & 0.6028 & 0.5589 & \underline{0.6162} & 0.5415 & 0.5943 & 0.5543 \\
            & {LPIPS$\downarrow$} & 0.3717 & 0.3925 & 0.3931 & 0.4631 & 0.4144 & 0.4189 & \textbf{0.3577} & \underline{0.3640} \\
            & {FID$\downarrow$} & 68.6481 & 66.2590 & 71.1709 & 55.1131 & 50.9362 & 57.7150 & \textbf{47.3327} & \underline{50.7819} \\
            & {DISTS$\downarrow$} & 0.2092 & 0.1858 & 0.2097 & 0.2363 & 0.2191 & \underline{0.1785} & 0.1959 & \textbf{0.1778} \\
            & {MUSIQ$\uparrow$} & 61.6423 & 60.0450 & 58.7220 & 56.3572 & 49.3028 & \textbf{66.9539} & \underline{66.3071} & 66.1278 \\
            & {NIQE$\downarrow$} & 3.6754 & 3.5621 & 3.6044 & 6.6093 & 4.4371 & \textbf{2.8572} & 3.9528 & \underline{3.3992} \\
            & {QAlign$\uparrow$} & 4.2460 & 3.6302 & 3.9947 & 4.1360 & 3.6496 & 4.3099 & \underline{4.3146} & \textbf{4.3175} \\
          \hline
           \multirow{8}{*}{RealSR} & PSNR$\uparrow$ & 25.8450 & 25.4330 & 26.0279 & \underline{26.2353} & 26.2057 & 24.9872 & \textbf{26.5952} & 25.9301 \\
           & {SSIM$\uparrow$} & 0.7734 & 0.7540 & \textbf{0.7802} & 0.7047 & \underline{0.7767} & 0.6812 & 0.7689 & 0.7105 \\
           & {LPIPS$\downarrow$} & 0.2729 & 0.2927 & \underline{0.2594} & 0.4594 & \textbf{0.2565} & 0.3633 & 0.2796 & 0.2806 \\
           & {FID$\downarrow$} & 67.0156 & 63.4422 & 64.1658 & 61.6524 & 109.1054 & \underline{55.1668} & 58.3206 & \textbf{47.3440} \\
           & {DISTS$\downarrow$} & 0.1685 & 0.1941 & 0.1609 & 0.2563 & \textbf{0.1584} & 0.1870 & 0.1859 & \underline{0.1604} \\
           & {MUSIQ$\uparrow$} & 59.6881 & 58.7741 & 59.6442 & 53.3410 & 60.7065 & \underline{65.5173} & 64.2653 & \textbf{65.5979} \\
           & {NIQE$\downarrow$} & 4.6781 & 4.7577 & 4.6453 & 7.4331 & 4.9309 & \textbf{3.6873} & 5.2673 & \underline{4.1886} \\
           & {QAlign$\uparrow$} & 3.9186 & 3.6895 & 3.8204 & 3.8199 & 3.8686 & \underline{4.0999} & 3.8884 & \textbf{4.1250} \\
          \hline
           \multirow{8}{*}{DRealSR} & PSNR$\uparrow$ & 27.9116 & 26.5869 & 27.8427 & 28.2573 & \textbf{29.3013} & 27.2030 & \underline{29.1033} & 29.0948 \\
           & {SSIM$\uparrow$} & 0.8249 & 0.7688 & 0.8209 & 0.7295 & \textbf{0.8462} & 0.7073 & \underline{0.8278} & 0.7937 \\
           & {LPIPS$\downarrow$} & 0.2818 & 0.3374 & 0.2838 & 0.5443 & \textbf{0.2724} & 0.4639 & \underline{0.2803} & 0.2893 \\
           & {FID$\downarrow$} & 23.1844 & 19.5815 & 24.6355 & 17.7943 & 17.6825 & 16.8249 & \underline{16.2228} & \textbf{14.2049} \\
           & {DISTS$\downarrow$} & 0.1464 & 0.1766 & 0.1461 & 0.2715 & \underline{0.1435} & 0.2107 & 0.1782 &  \textbf{0.1429} \\
           & {MUSIQ$\uparrow$} & \textbf{35.2563} & 31.7808 & 34.6197 & 26.4684 & 33.1994 & 33.3549 & 30.5885 & \underline{34.6331} \\
           & {NIQE$\downarrow$} & 4.7146 & 4.2176 & 4.5669 & 7.1426 & 5.4360 & \textbf{2.9683} & 5.3846 & \underline{4.1390} \\
           & {QAlign$\uparrow$} & \underline{4.3029} & 4.2651 & 4.2824 & 4.2366 & 4.2427 & 4.2492 & 4.2178 & \textbf{4.3822}
       \end{tabular}}
    \label{tab:realisr}
    \vspace*{-3mm}
\end{table*}

\begin{figure*}[t!]
    \centering
    \includegraphics[width=0.96\textwidth]{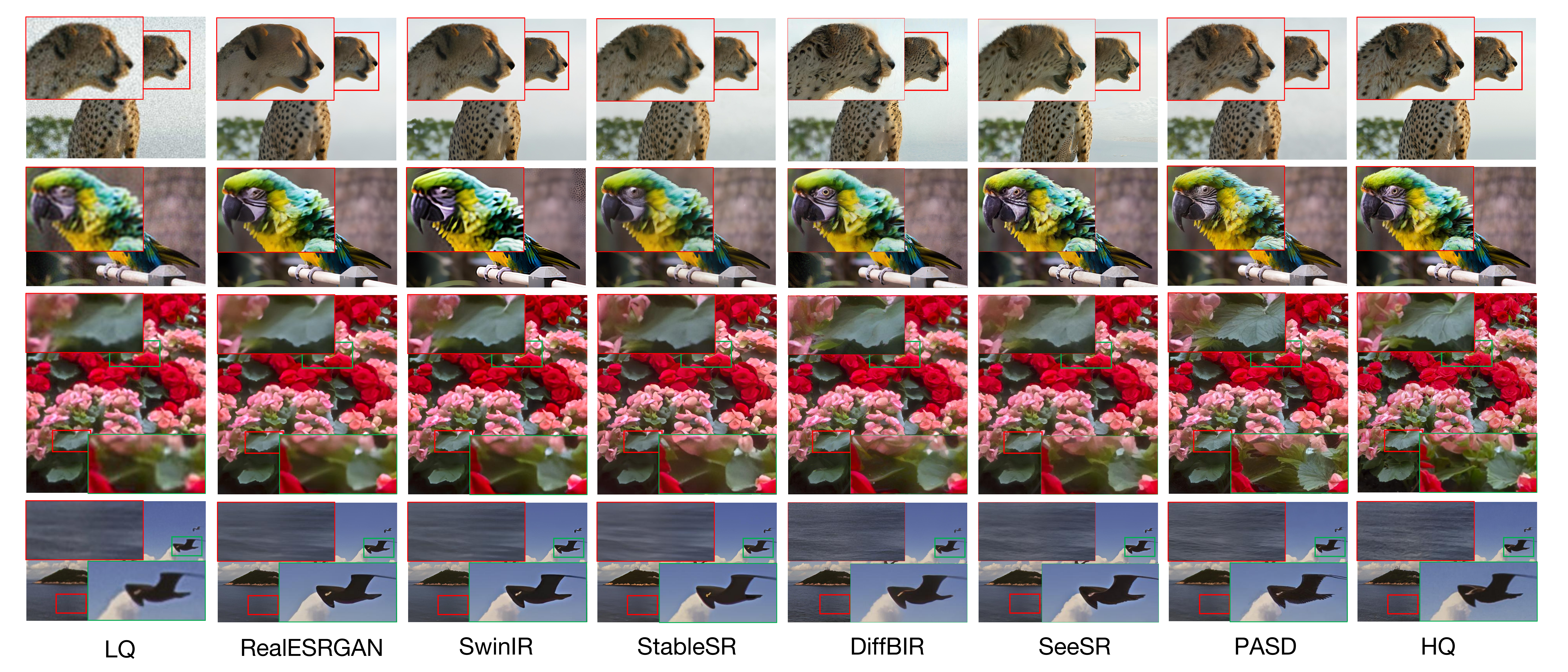}
   \vspace*{-2mm}
    \caption{Realistic image super-resolution results by different methods.}
    \label{fig:realisr}
    \vspace{-2mm}
\end{figure*}

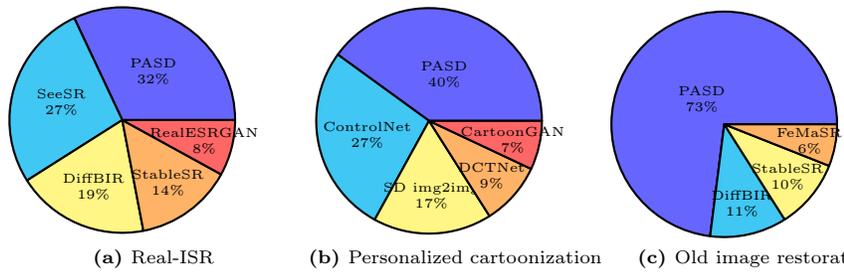
\begin{figure}[t!]
\centering
    \begin{subfigure}{0.32\textwidth}
        \begin{tikzpicture}[scale=0.5]
            \pie[radius=3,text=inside,font=\tiny]{32/PASD, 27/SeeSR, 19/DiffBIR, 14/StableSR, 8/RealESRGAN}
        \end{tikzpicture}
        \caption{Real-ISR}
        \label{fig:ustudy1}
    \end{subfigure}
    \begin{subfigure}{0.32\textwidth}
        \begin{tikzpicture}[scale=0.5]
            \pie[radius=3,text=inside,font=\tiny]{40/PASD, 27/ControlNet, 17/SD img2img, 9/DCTNet, 7/CartoonGAN}
        \end{tikzpicture}
        \caption{Personalized cartoonization}
        \label{fig:ustudy2}
    \end{subfigure}
    \begin{subfigure}{0.32\textwidth}
        \begin{tikzpicture}[scale=0.5]
            \pie[radius=3,text=inside,font=\tiny]{73/PASD, 11/DiffBIR, 10/StableSR, 6/FeMaSR }
        \end{tikzpicture}
        \caption{Old image restoration}
        \label{fig:ustudy3}
    \end{subfigure}
    \vspace{-2mm}
    \caption{User study results of (a) Real-ISR, (b) personalized cartoonization, and (c) old photo restoration tasks.}
    \vspace{-2mm}
    \label{fig:ustudy}
\end{figure}

\begin{table*}[t!]
    \centering
    \caption{The FID, MUSIQ and QAlign indices of different cartoonization models on test data. The best and second best scores are in bold and underscore.}
    \vspace*{-2mm}
        \normalsize{
        \resizebox{0.96\textwidth}{!}{\begin{tabular}{c|c|c c c|c c c|c}
            Datasets & Metrics & CartoonGAN & AnimeGAN & DCTNet & InstructPix2Pix & SD Img2img & ControlNet & PASD \\
          \hline\hline
            \multirow{3}{*}{FFHQ} & {FID$\downarrow$} & 53.7454 & 58.4010 & 50.6869 & 39.3259 & 63.0608 & \underline{37.9589} & \textbf{37.6698} \\
            & {MUSIQ$\uparrow$} & 71.9848 & 70.3372 & 62.6950 & 72.9510 & \underline{74.8963} & 74.7521 & \textbf{75.0221} \\
            & {QAlign$\uparrow$} & 3.6064 & 3.4988 & 3.6812 & 3.9030 & \textbf{4.0076} & 3.9233 & \underline{3.9978} \\
          \hline
           \multirow{3}{*}{Flicker2K} & {FID$\downarrow$} & 72.5560 & 78.4741 & 81.0789 & \underline{71.2098} & 75.5553 & 72.2742 & \textbf{70.3800} \\
           & {MUSIQ$\uparrow$} & 71.5768 & 72.7070 & 72.2595 & \underline{74.1627} & \textbf{75.9558} & 72.3908 & 73.1320 \\
           & {QAlign$\uparrow$} & 3.7277 & 3.6835 & 3.8483 & 3.8842 & 3.9008 & \textbf{3.9331} & \underline{3.9290} \\
       \end{tabular}}}
    \label{tab:ps}
    \vspace*{-3mm}
\end{table*}

\begin{figure*}[t!]
    \centering
    \includegraphics[width=0.96\textwidth]{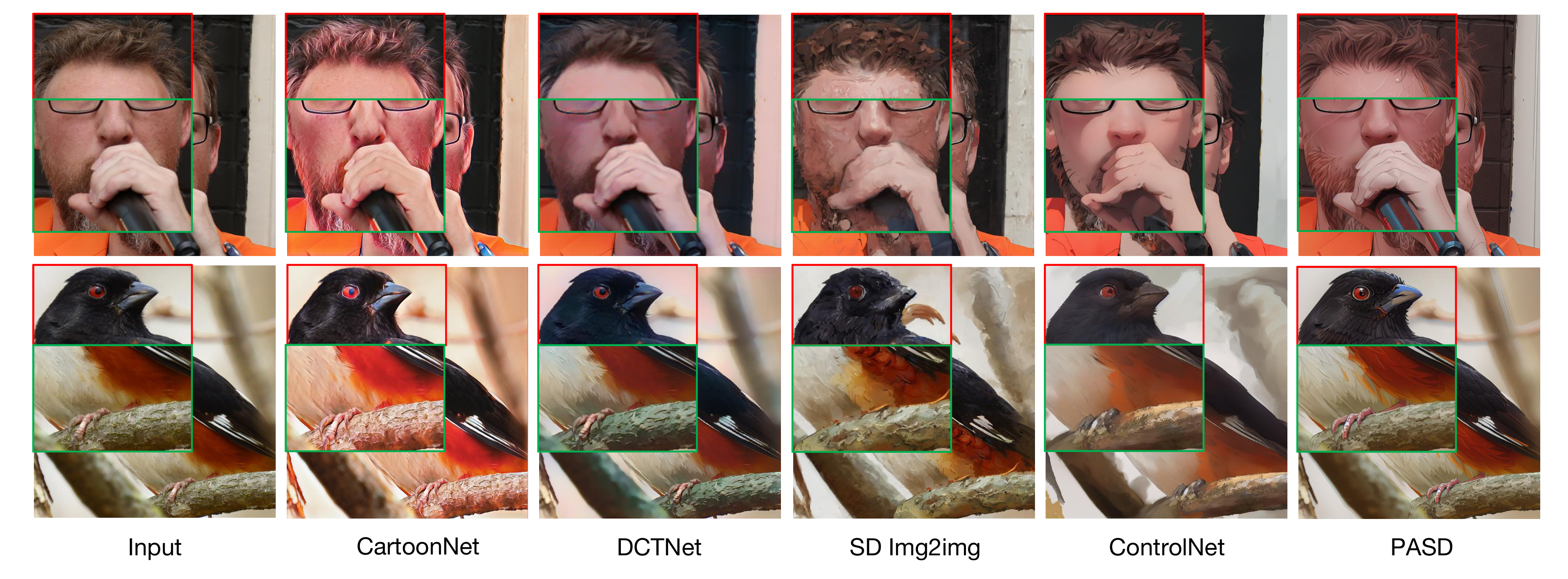}
    \vspace*{-3mm}
    \caption{Stylization (cartoonization) results by different methods on real-world images.}
    \label{fig:ps}
    \vspace*{-3mm}
\end{figure*}

\begin{figure*}[t!]
    \centering
    \includegraphics[width=0.96\textwidth]{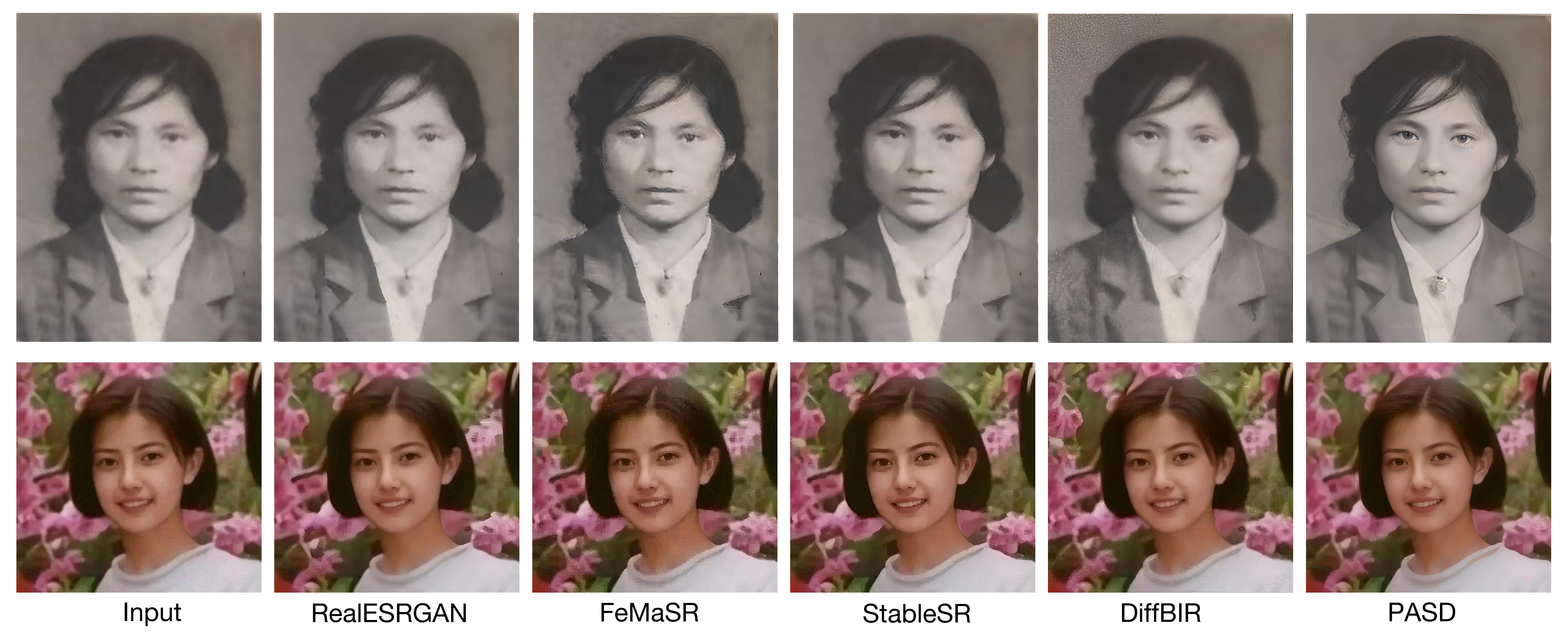}
    \vspace*{-3mm}
    \caption{Old image restoration results by different methods on real-world images.}
    \label{fig:old}
    \vspace*{-3mm}
\end{figure*}

\begin{table*}[t!]
    \centering
    \caption{The FID, MUSIQ and QAlign indices of different old image restoration models on self-collected data. The best and second best scores are in bold and underscore.}
   \vspace*{-2mm}
        \small{
        \resizebox{0.9\textwidth}{!}{\begin{tabular}{c|c c c c|c c|c}
             Metrics & RealESRGAN & FeMaSR & SwinIR & Wan \etal & StableSR & DiffBIR & PASD \\
          \hline\hline
            {FID$\downarrow$} & 265.5364 & 263.5393 & 266.9950 & 268.3455 & 264.3178 & \underline{262.1837} & \textbf{240.2576} \\
            {MUSIQ$\uparrow$} & 59.4676 & \underline{61.2766} & 55.3312 & 32.1470 & 53.2194 & 60.3904 & \textbf{64.4023} \\
            {QAlign$\uparrow$} & 3.6660 & 3.6347 & 3.8232 & 3.0120 & 3.8636 & \underline{3.9191} & \textbf{3.9797} \\
       \end{tabular}}}
    \label{tab:old}
    \vspace*{-3mm}
\end{table*}


\subsection{Effectiveness of the Adjustable Noise Schedule}
We first use the task of Real-ISR to discuss the setting and advantages of our proposed ANS. In order to find out how $\bar{\alpha}_a$ affects the performance, we set $\bar{\alpha}_a\in\{0,0.1,0.5,1\}$) and perform experiments on the RealSR test dataset \cite{cai2019realsr}. The curves of PSNR/QAlign versus $\bar{\alpha}_a$ are plotted in Fig.~\ref{fig:curve}. One can see that the PSNR performance increases while the QAlign score decreases as $\bar{\alpha}_a$ grows, demonstrating that the proposed $\bar{\alpha}_a$ can be employed to enable flexible perception-fidelity trade-off. Fig.~\ref{fig:ans} visualizes Real-ISR results with different values of $\bar{\alpha}_a$. We can see that with the increase of $\bar{\alpha}_a$, PASD tends to improve the fidelity while generate less realistic details. In practice, we choose $\bar{\alpha}_a$ from the values of $\bar{\alpha}_n$, where $n\in\{1,2...N\}$, for convenience. In all of our following experiments, we empirically set $n=900$, \ie, $\bar{\alpha}_{900}=0.1189$, to achieve a good balance between fidelity and perception quality. 

\vspace{-2mm}
\subsection{Experimental Results}
\textbf{Realistic image super-resolution.}
We compare the proposed PASD method with two categories of Real-ISR algorithms. The first category is GAN-based methods, including Real-ESRGAN \cite{wang2021realesrgan}, FeMaSR \cite{chen2022femasr}, and SwinIR \cite{liang2021swinir}. The second category is diffusion-based models, including ResShift \cite{yue2023resshift}, StableSR \cite{wang2023stablesr}, DiffBIR \cite{lin2023diffbir}, and SeeSR \cite{wu2023seesr}. The quantitative evaluation results on the test data are presented in Tab.~\ref{tab:realisr}, from which we can have the following observations.

First, in term of fidelity measures PSNR/SSIM, the diffusion-based methods are not advantageous over GAN-based methods. This is because diffusion models have higher generative capability and hence may synthesize more perceptually realistic but less faithful details, resulting in lower PSNR/SSIM indices. Second, the diffusion-based methods, especially the proposed PASD, perform better than GAN-based methods in most perception metrics. This conforms to our observation on the visual quality (see Fig.~\ref{fig:realisr}) of their Real-ISR output. Third, PASD achieves the best QAlign scores, which is a no-reference image quality assessment index based on large vision-language models, on all the three test datasets.

Fig.~\ref{fig:realisr} visualizes the Real-ISR results of competing methods. It can be seen that our PASD method can generate more realistic details with better visual quality (see the synthesized textures in fur, flowers, leaves, feathers, sea, etc.). Fig.~\ref{fig:ustudy1} presents the results of subjective user study. PASD receives the most rank-1 votes, confirming its superiority in generating realistic image details. More visual results can be found in the \textbf{supplementary material}.

\textbf{Personalized cartoonization.}
Similar to the Real-ISR task, we compare the proposed PASD with two categories of stylization algorithms. The first category is GAN-based methods, including CartoonGAN \cite{chen2018cartoongan}, AnimeGAN \cite{chen2020animegan} and DCTNet \cite{men2022dctnet}. We re-train these models with a batch of stylized images generated by a personalized diffusion model, \ie, ToonYou \footnote{\url{https://civitai.com/models/30240/toonyou}}. The second category is diffusion-based algorithms, including InstructPix2Pix \cite{brooks2022instructpix2pix}, SD img2img \cite{rombach2021latent} and ControlNet \cite{zhang2023controlnet}. We replace their base models with the personalized model for fair comparison. 
Tab.~\ref{tab:ps} shows the quantitative evaluation results. It can be seen that PASD achieves the best or second best results in most indices.

Fig.~\ref{fig:ps} shows some cartoonization results. One can see that compared with GAN-based methods, the results of PASD is much cleaner. Compared with the diffusion-based models, PASD can better preserve image details such as human hair. Due to the limited space, we only present results with the style of ToonYou here. Please note that PASD can generate various stylization results by simply switching the base diffusion model to a personalized one without any additional training procedure. More stylization results, including the results on image colorization, can be found in the \textbf{supplementary materials}.

As in the task of Real-ISR, we also conducted a user study for subjective assessment on the image stylization performance. Fig.~\ref{fig:ustudy2} shows the results. Clearly, PASD is preferred by most subjects. 


\textbf{Old photo restoration.}
We compare PASD with Wan \etal \cite{wan2021bringing} and several real-world SR methods, including RealESRGAN \cite{wang2021realesrgan}, FeMaSR \cite{chen2022femasr}, SwinIR \cite{liang2021swinir}, StableSR \cite{wang2023stablesr}, and DiffBIR \cite{lin2023diffbir}. We re-use the PASD model trained for Real-ISR task but replace its base model with an aesthetic one ,\ie majicMIX realistic \footnote{\url{https://civitai.com/models/43331/majicmix-realistic}}.

Tab.~\ref{tab:old} shows the quantitative evaluation results. It can be seen that PASD achieves the best results in all three indices. 
Fig.~\ref{fig:old} visualizes some old photo restoration results. Compared with the competing methods, PASD can better recover vivid image details such as human hair. Fig.~\ref{fig:ustudy3} presents the results of subjective user study. Clearly, PASD is preferred by the majority of subjects. More visual results can be found in the \textbf{supplementary material}.

\begin{figure*}[t!]
    \centering
    \includegraphics[width=0.96\textwidth]{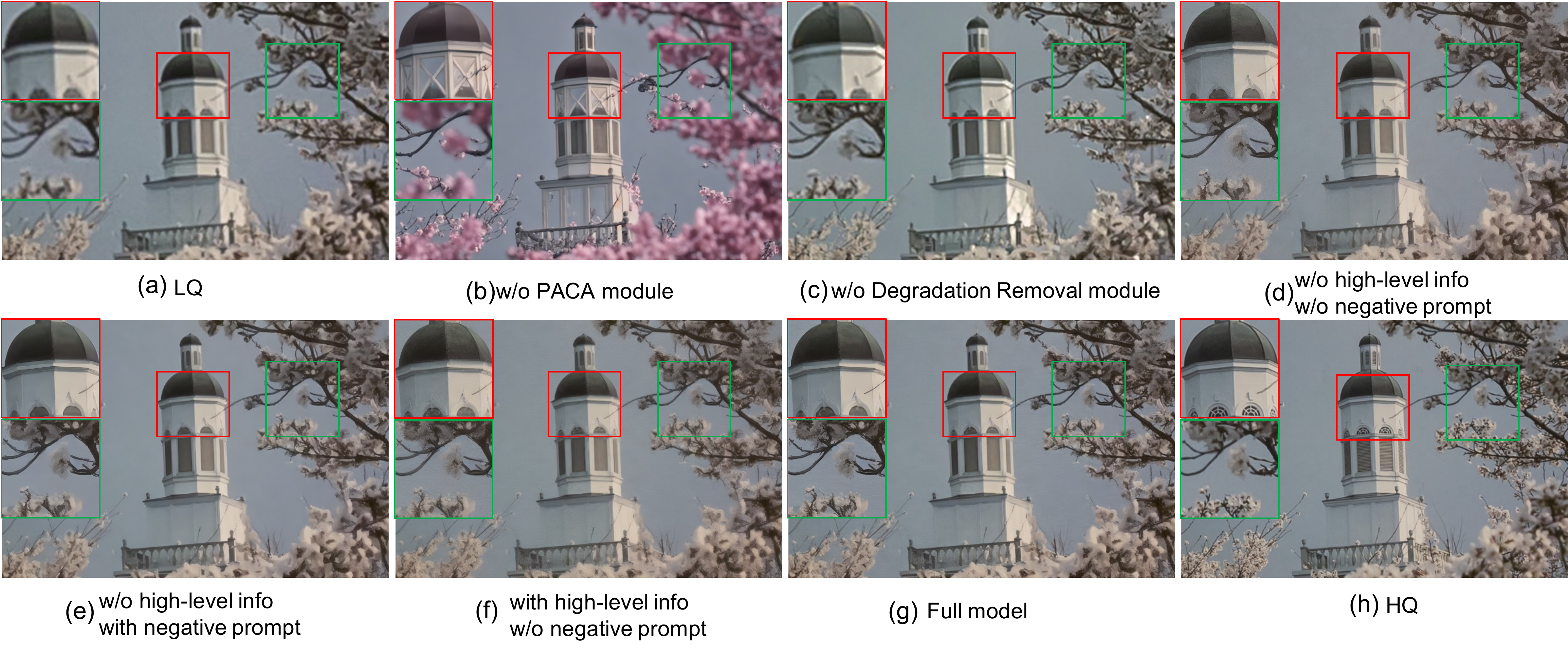}
    \vspace*{-2mm}
    \caption{Real-ISR results by different variants of PASD.}
    \vspace{-1mm}
    \label{fig:ablation}
\end{figure*}

\begin{table}[t!]
    \caption{Quantitative results of different variants of PASD on RealSR test dataset.}
    \centering
    \vspace*{-2mm}
    \resizebox{0.96\textwidth}{!}{\begin{tabular}{l|c c c|c c c|c}
        Exp. & \makecell{Degradation Removal} & \makecell{High-level info} & \makecell{Negative prompt} & PSNR$\uparrow$ & FID$\downarrow$ & LPIPS$\downarrow$ & Runtime(s)$\downarrow$\\
        \hline
        (a) & & \checkmark & \checkmark & 26.1108 & 56.7890 & 0.3822 & 14.3221 \\
        (b) & \checkmark & & & \textbf{27.8731} & 53.8988 & 0.3080 & \textbf{8.0368} \\
        (c) & \checkmark & \checkmark & & 27.0901 & 52.3380 & 0.2851 & \underline{8.7366} \\
        (d) & \checkmark & & \checkmark & \underline{27.3781} & \underline{50.2502} & \underline{0.2809} & 13.3163 \\
        (e) & \checkmark & \checkmark & \checkmark & 25.9301 & \textbf{47.3440} & \textbf{0.2806} & 14.5889\\
    \end{tabular}}
    \vspace*{-3mm}
\label{tab:ablation}
\end{table}

\vspace{-2mm}
\subsection{Ablation Studies}
We perform a series of ablation studies of the proposed PASD network, including the importance of PACA, the role of degradation removal module, and the role of high-level information. We visualize the Real-ISR results of different variants of PASD in Fig.~\ref{fig:ablation}, and report the quantitative results and runtime in Tab.~\ref{tab:ablation}. 

\textbf{Importance of PACA.}
We evaluate a variant of PASD by excluding the PACA module from it, \ie, the features $\mathbf{y}$ extracted from ControlNet are simply added to features $\mathbf{x}$. As shown in Fig.~\ref{fig:ablation}(b), the output becomes inconsistent with the LQ input in colors and structures, etc. This verifies the importance of PACA in perceiving pixel-wise local structures.

\textbf{Role of degradation removal module.}
To evaluate the effect of degradation removal module, we remove the ``toRGB'' modules as well as the pyramid $\mathcal{L}_{DR}$ loss during model training. As can be seen in Fig.~\ref{fig:ablation}(c) and Tab.~\ref{tab:ablation}, removing the degradation removal module leads to dirty outputs and worse PSNR, FID and LPIPS indices.

\textbf{Role of high-level information.}
The high-level information and negative prompt are optional but very useful for PASD. We simply replace them with null-text prompt to evaluate their effects. As shown in Fig.~\ref{fig:ablation}(d), replacing both high-level information and negative prompt with null-text prompt results in dirty outputs with less realistic details, which is also verified by the worse FID and LPIPS indices in Tab.~\ref{tab:ablation}. Abandoning high-level information leads to over-smoothed results, as illustrated in Fig.~\ref{fig:ablation}(e). The output can become dirty without negative prompt (see Fig.~\ref{fig:ablation}(f)). Our full model takes advantages of both high-level information and negative prompt, and achieves a good balance between clean-smooth and detailed-dirty outputs (see Fig.~\ref{fig:ablation}(g) and the best FID score in Tab.~\ref{tab:ablation}).

\textbf{Runtime analysis on different modules.}
The runtime is reported as the average over $10$ runs to process a $256\times 256$ image on a NVIDIA Tesla 32G-V100 GPU. We use the DDIM \cite{song2021ddim} sampler for $20$ steps.
By comparing Exps. (a) and (e) in Tab.~\ref{tab:ablation}, one can see that the degradation removal module has little effect on the runtime. Without the negative prompt module, the runtime nearly cuts in half because the classifier-free guidance can be removed (see Exps.(c) and (e)). Finally, the high-level information module only increase a little the runtime  (see Exps.(d) and (e)).

\vspace{-1mm}
\section{Conclusion and Limitation}
\vspace{-1mm}
We proposed a pixel-aware diffusion network, namely PASD, for realistic image restoration and personalized stylization. By introducing a pixel-aware cross attention module, PASD succeeded in perceiving image local structures in pixel-level and achieved robust and perceptually realistic Real-ISR results. An adjustable noise schedule was also proposed, which helped PASD to achieve flexible perception-fidelity trade-off during the inference stage. By replacing the base model to a personalized one, PASD could produce diverse stylization results with highly consistent semantic contents with the input. The proposed PASD was simple to implement, and our extensive experiments demonstrated its effectiveness and flexibility across different tasks, showing its great potentials for handling complex image restoration and stylization tasks.

Though PASD can achieve pixel-level enhancement, it still suffers from the balance between fidelity and perception. In addition, it may fail to reproduce faithful details when the input image is heavily degraded or the semantic information is inaccurate. A more robust degradation estimation module can be designed, and more precise semantic information can be extracted to further improve the performance of PASD, which will be considered in our future work.

%
%
\bibliographystyle{splncs04}
\bibliography{arxiv}

\clearpage

\appendix

\section*{\centering Supplementary Materials} 

In this supplementary file, we provide the following materials:
\begin{itemize}
    \setlength{\itemsep}{0.7pt}
    \setlength{\parsep}{0.7pt}
    \setlength{\parskip}{0.7pt}
    \item More visual comparisons of different Real-ISR models;
    \item Various stylization results generated by PASD;
    \item More visual comparisons of different old image restoration methods;
    \item Results of PASD on image colorization.
\end{itemize}

\subsection*{More Real-ISR Results}
In Fig.~\ref{fig:realisr}, we show more visual comparisons between our method with state-of-the-art Real-ISR methods, including RealESRGAN \cite{wang2021realesrgan}, SwinIR \cite{liang2021swinir}, StableSR \cite{wang2023stablesr}, DiffBIR \cite{lin2023diffbir} and SeeSR \cite{wu2023seesr}. Similar conclusions to the main paper can be made. With the help of PACA and ANS modules, our PASD can provide adjustable pixel-level guidance on the image generation, reproducing more realistic fine details and less visual artifacts.

\subsection*{Various Stylization Results}
As mentioned in the main paper, by simply switching the base diffusion model to a personalized one, our proposed PASD can do various stylization tasks without any additional training procedure. In the main paper, we have provided the results by using the ToonYou style. In Fig~\ref{fig:ps} of this supplementary file, we show more types of stylization results by using the personalized base models of Disney 3D, Oil painting and Shinkai. One can see that our PASD method can keep very well the pixel-wise image details while performing style transfer.  

\subsection*{More Old Image Restoration Results}
In Fig.~\ref{fig:old}, we show more visual comparisons between our method with state-of-the-art old image restoration methods, including Real-ESRGAN \cite{wang2021realesrgan}, FeMaSR \cite{chen2022femasr}, StableSR \cite{wang2023stablesr}, and DiffBIR \cite{lin2023diffbir}. It can be seen that PASD can better recover semantic-aware and photo-realistic image details.

\subsection*{Image Colorization}
Our PASD can serve as a generic solution for various pixel-wise image-to-image tasks. In addition to Real-ISR, personalized stylization and old photo restoration, we also apply it to image colorization and show the results in this supplementary file. 

The proposed ANS in the main paper can effectively solve the non-zero SNR issue without any extra training in tasks such as Real-ISR. However, it cannot be directly applied to the task of image colorization, where the residual signal from RGB image cannot be compensated from the input grayscale image. To solve this problem, we follow the idea proposed in \cite{lin2023snr} to scale the noise schedule as follows:
\begin{equation}
\sqrt{\bar{\alpha'}_t}=\frac{\sqrt{\bar{\alpha}_t}-\sqrt{\bar{\alpha}_N}}{\sqrt{\bar{\alpha}_1}-\sqrt{\bar{\alpha}_N}},
\end{equation}
where $t\in\{1,2...N\}$ and $\alpha'$ is the scaled $\alpha$. One can see that $\sqrt{\bar{\alpha'}_{N-1}}\neq0$ and $\sqrt{\bar{\alpha'}}_N=0$, which means that $\alpha'_N$ has been successfully set to $0$.

When SNR is zero, the $\epsilon$ prediction used in SD becomes a trivial task because the $\epsilon$ loss cannot guide the model to learn meaningful information from the data. We therefore follow \cite{salimans2022v-predict,lin2023snr} to finetune the PASD model with rescaled noise schedule and $\nu$ loss and prediction.

\begin{figure*}[t!]
    \centering
    \includegraphics[width=0.96\textwidth,height=16cm]{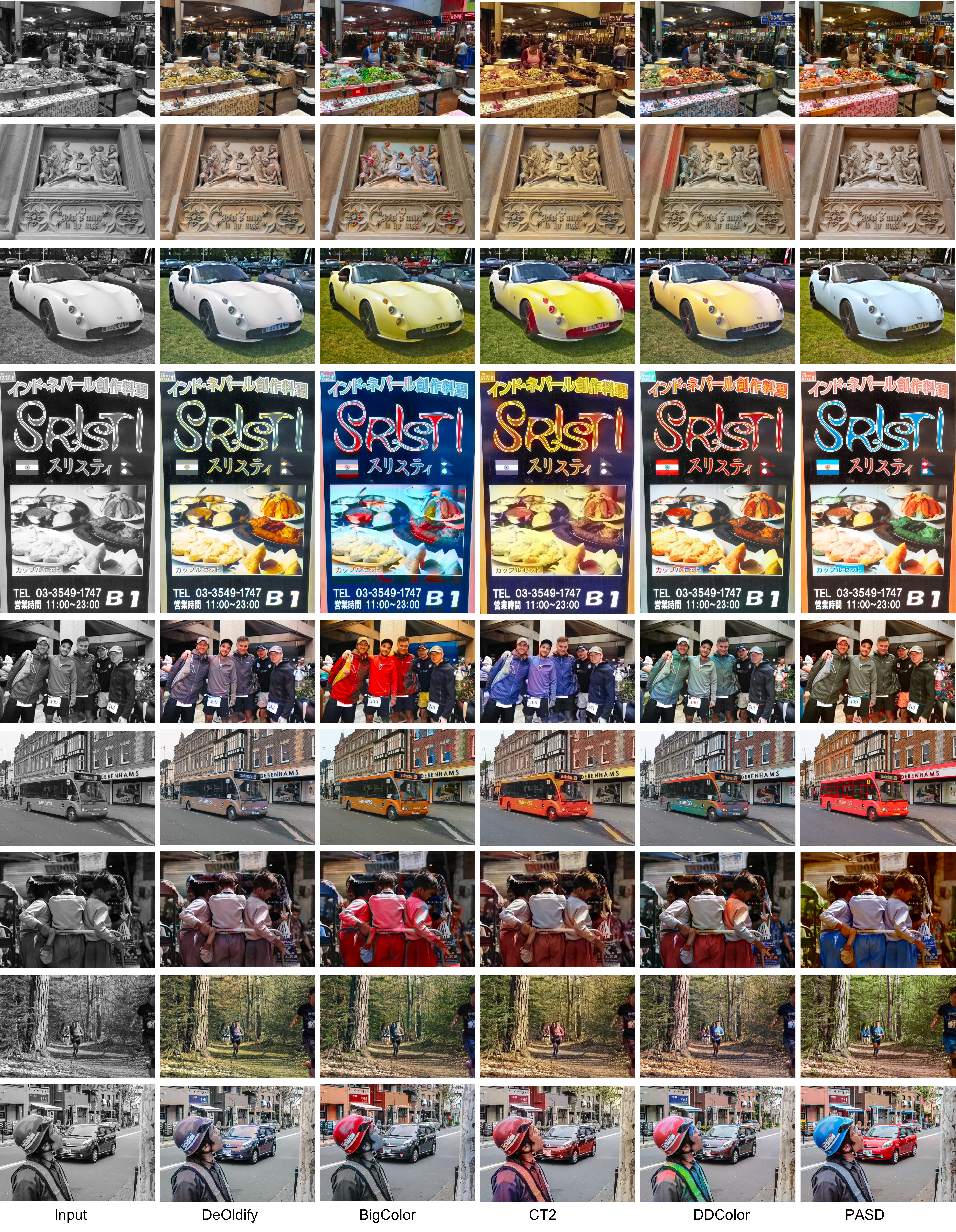}
    \caption{Qualitative comparison of different colorization methods on real-world images.}
\label{fig:color}
\end{figure*}

Figure~\ref{fig:color} shows the qualitative comparisons between PASD and the state-of-the-art image colorization methods, including DeOldify \cite{anti2019deoldify}, BigColor \cite{kim2022bigcolor}, CT2 \cite{weng2022ct2} and DDColor \cite{kang2023ddcolor}. One can see that our PASD generates more photo-realistic and vivid colorization results. In particular, it significantly alleviates the color bleeding effect, which often happens in the compared methods.

\begin{figure*}[t!]
    \centering
    \includegraphics[width=0.96\textwidth,height=14.5cm]{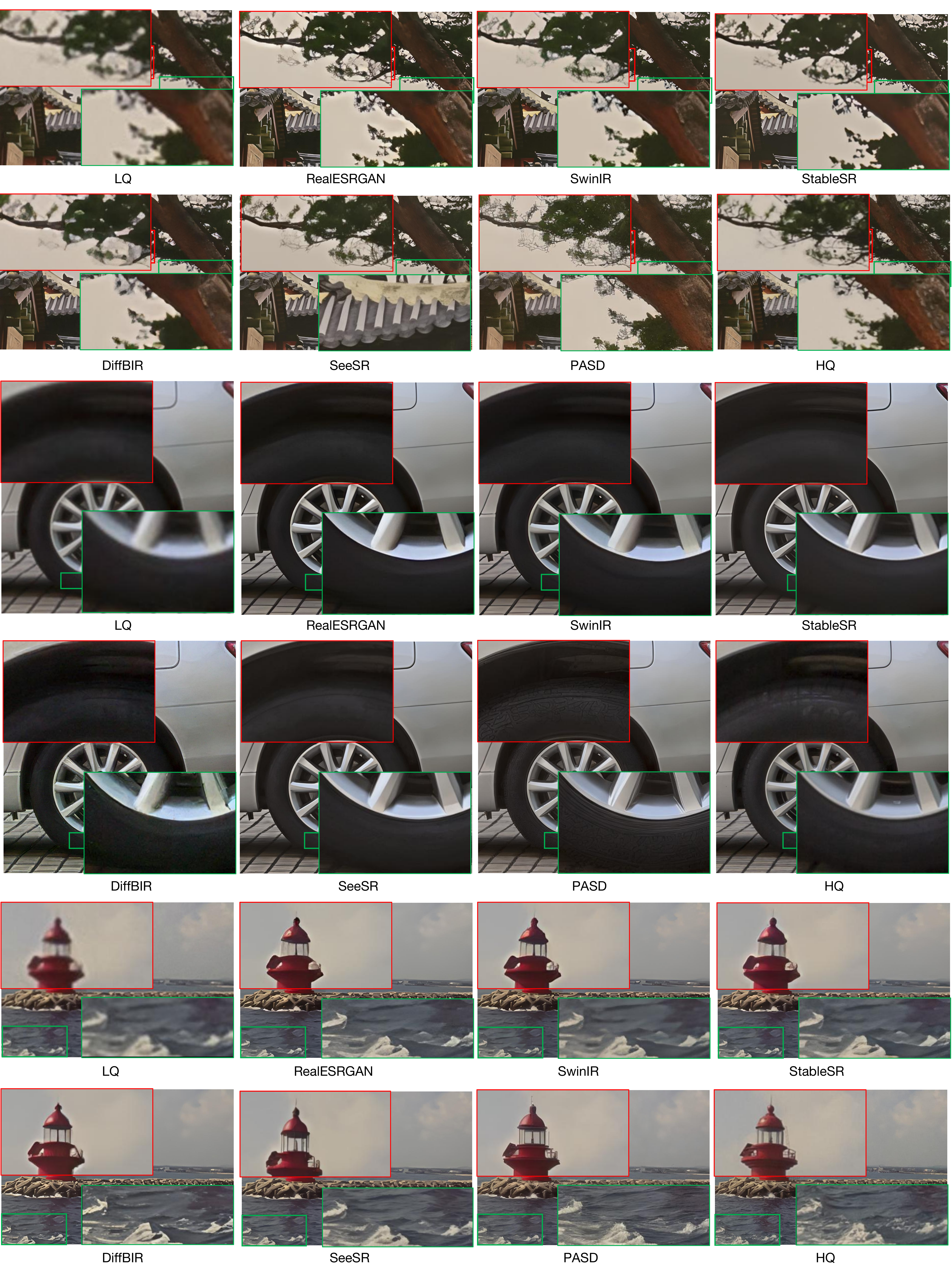}
    \caption{Realistic image super-resolution results by different methods. Please zoom-in for better comparison.}
    \vspace{-3mm}
\label{fig:realisr}
\end{figure*}

\begin{figure*}[t!]
    \centering
    \includegraphics[width=0.96\textwidth]{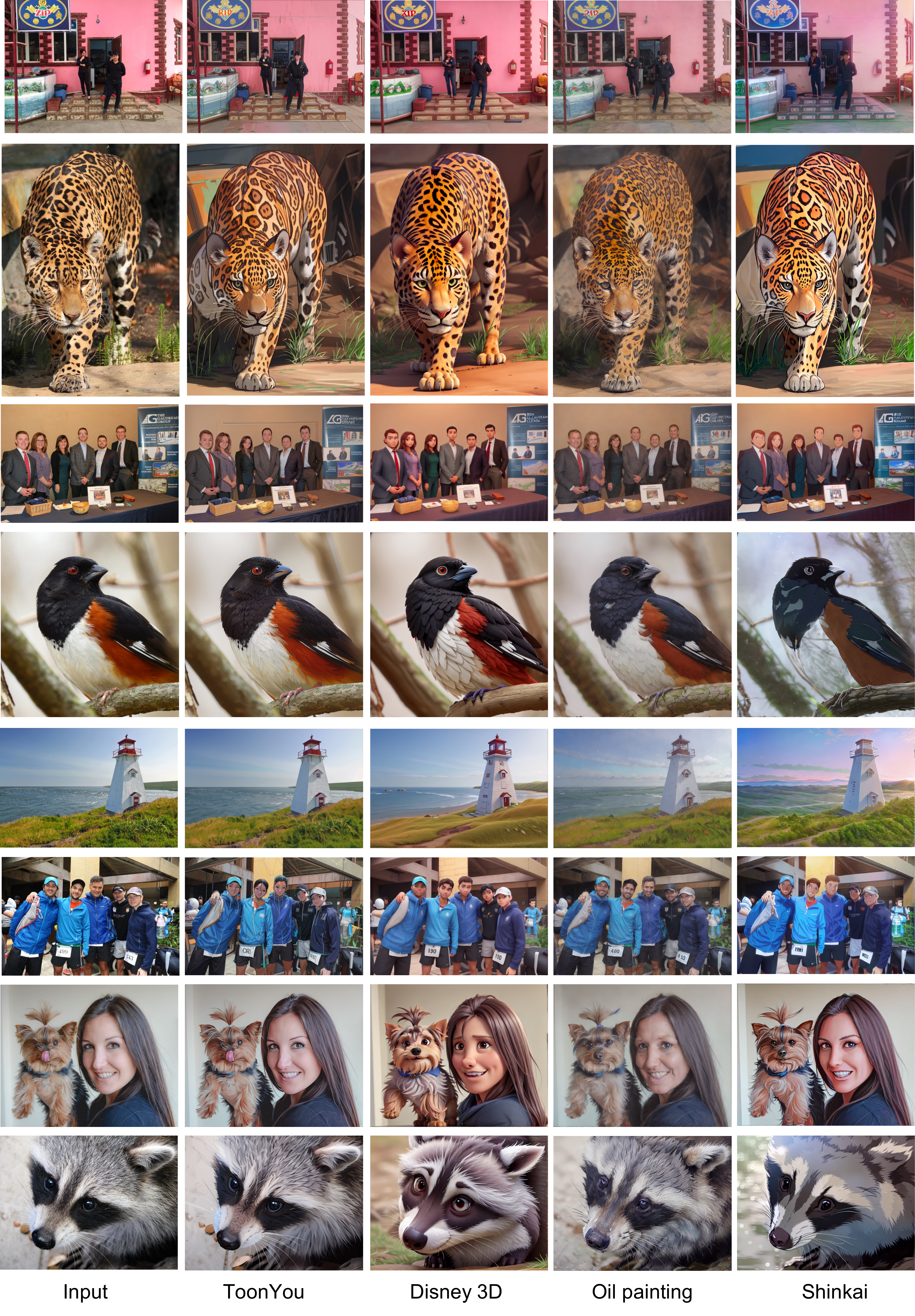}
    \caption{Stylization results by PASD with different base models (ToonYou, Disney 3D, Oil painting, Shinkai) on real-world images.}
\label{fig:ps}
\end{figure*}

\begin{figure*}[t!]
    \centering
    \includegraphics[width=0.96\textwidth,height=16cm]{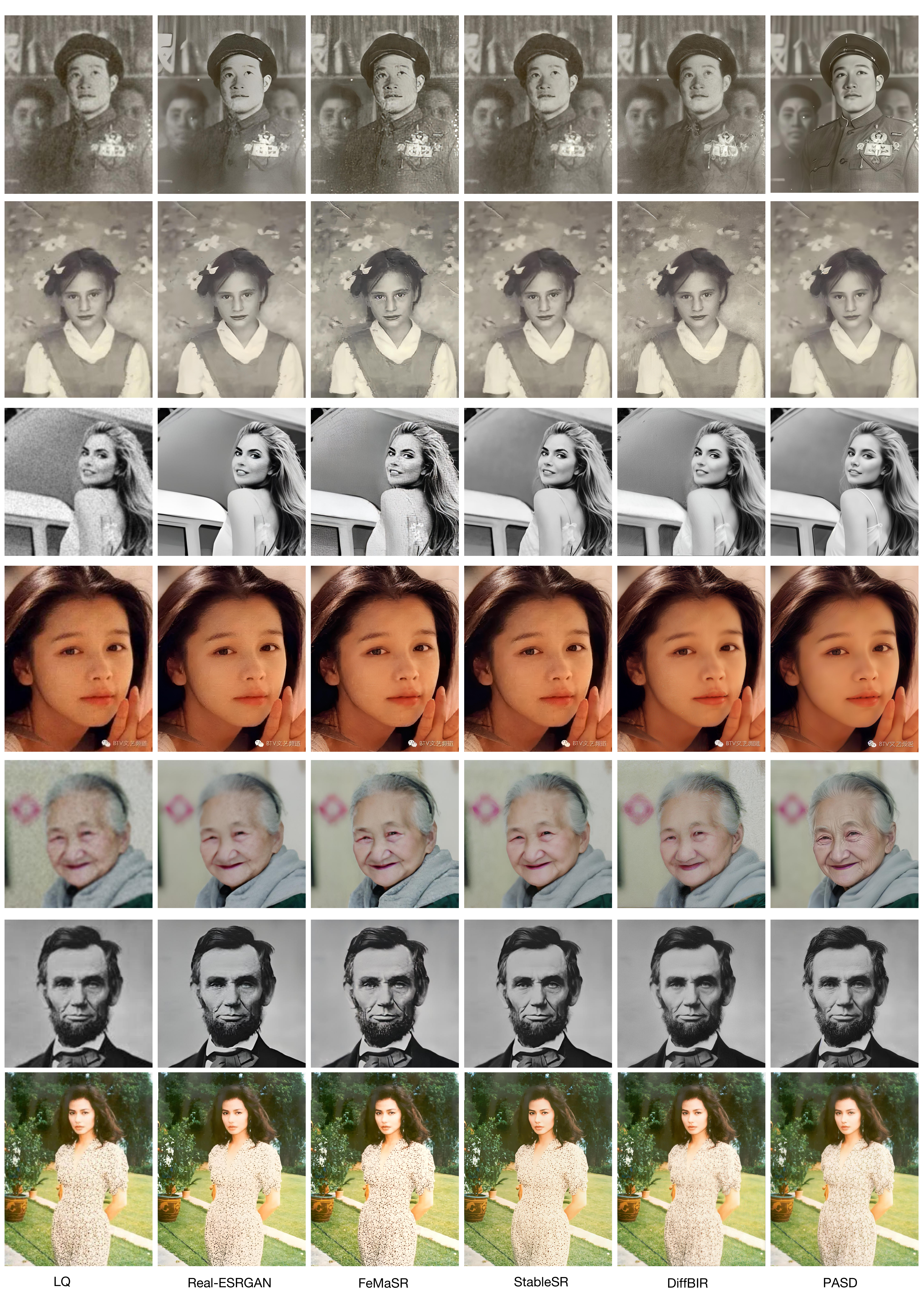}
    \caption{Old image restoration results by different methods. Please zoom-in for better comparison.}
\label{fig:old}
\end{figure*}

\end{document}